\documentclass[10pt, conference, compsocconf]{IEEEtran}
\ifCLASSINFOpdf
\else
\fi
\usepackage{amssymb}
\usepackage{amsmath}
\usepackage{graphicx}
\usepackage{subcaption}
\usepackage{pgfplots}
\usepackage{cite}
\usepackage{hyperref}
\usepackage{array}
\usepackage{xcolor}
\hypersetup{
    colorlinks=true,
    linkcolor=blue,
    filecolor=magenta,      
    urlcolor=cyan,
}
\usepackage{acronym}

\acrodef{VPR}{visual place recognition}
\acrodef{CNN}{convolutional neural network}
\acrodef{PCA}{principal component analysis}

\begin{document}
\title{Compact Environment-Invariant Codes for Robust Visual Place Recognition}


\author{\IEEEauthorblockN{Unnat Jain}
\IEEEauthorblockA{Dept. of Computer Science\\
University of Illinois Urbana-Champaign\\
Urbana, USA\\
uj2@illinois.edu}
\and
\IEEEauthorblockN{Vinay P. Namboodiri}
\IEEEauthorblockA{Dept. of Computer Science \& Engg.\\
Indian Institute of Technology\\
Kanpur, India\\
vinaypn@iitk.ac.in}
\and
\IEEEauthorblockN{Gaurav Pandey}
\IEEEauthorblockA{Dept. of Electrical Engg.\\
Indian Institute of Technology\\
Kanpur, India\\
gpandey@iitk.ac.in}
}
% \thanks{*This work is supported by Ford Motor Company through the project FMT/EE/2015241}% <-this % stops a space

%Telephone: (800) 555--1212, Fax: (888) 555--1212}


\maketitle
\begin{abstract}
Robust \ac{VPR} requires scene representations that are invariant to various environmental challenges such as seasonal changes and variations due to ambient lighting conditions during day and night. Moreover, a practical VPR system necessitate compact representations of environmental features. To satisfy these requirements, in this paper we suggest a modification to the existing pipeline of \ac{VPR} systems to incorporate supervised hashing. The modified system learns (in a supervised setting) compact binary codes from image feature descriptors. These binary codes imbibe robustness to the visual variations exposed to it during the training phase, thereby, making the system adaptive to severe environmental changes. Also, incorporating supervised hashing makes VPR computationally more efficient and easy to implement on simple hardware. This is because binary embeddings can be learnt over simple-to-compute features and the distance computation is also in the low dimensional hamming space of binary codes. We have performed experiments on several challenging data sets covering seasonal, illumination and viewpoint variations. We also compare two widely used supervised hashing methods of CCAITQ \cite{itq} and MLH \cite{mlh} and show that this new pipeline out-performs or closely matches the state-of-the-art deep learning \ac{VPR} methods that are based on high-dimensional features extracted from pre-trained deep convolutional neural networks.

% \vpn{We do not propose a new algorithm in this paper, rather we evaluate existing algorithms, we might be overclaiming which may annoy reviewers. This can be clarified. Suggested sentence, some variant of the following: We modify the existing pipeline of \ac{VPR} systems to incorporate supervised hashing. We evaluate several existing supervised hashing methods show that this new pipeline outperforms or closely matches the state of the art deep learning methods...}
\end{abstract}

\begin{IEEEkeywords}
visual place recognition; similarity learning; hashing; convolutional neural network; dynamic time warping

\end{IEEEkeywords}

\IEEEpeerreviewmaketitle

\section{Introduction}
\begin{figure}[t]
    \centering
    \begin{subfigure}[c]{1\linewidth}
        \centering
        \includegraphics[width=\linewidth]{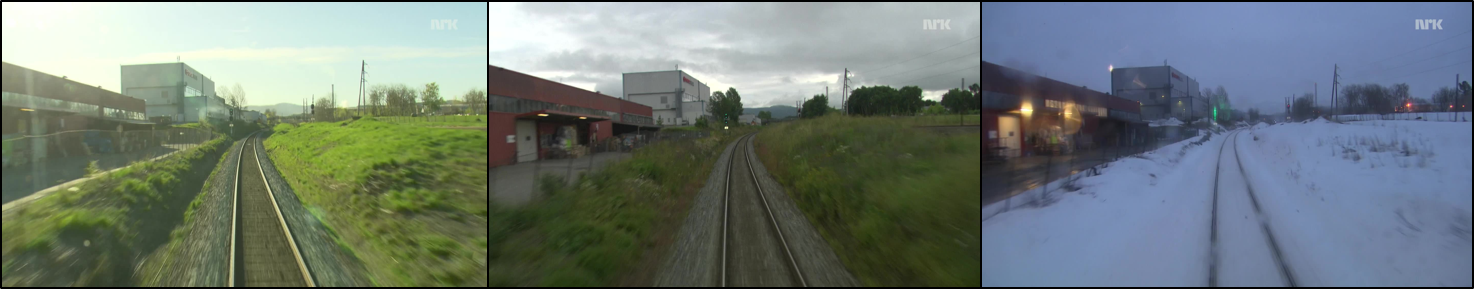}
        \caption{Nordland spring-summer \& summer-winter data sets: Mild to severe appearance change \& no viewpoint change.}    
        \label{fig:datasets_1}
    \end{subfigure}
    
    \vskip\baselineskip
    \begin{subfigure}[c]{1\linewidth}   
        \centering 
        \includegraphics[width=\linewidth]{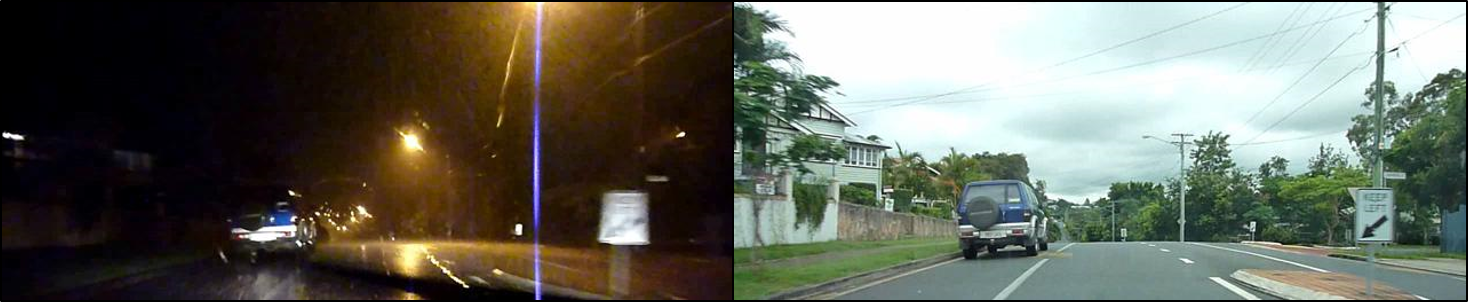}
        \caption{Alderley Day-Night data set: Severe appearance \& mild viewpoint change.}  
        \label{fig:datasets_2}
        % \label{fig:mean and std of net34}
    \end{subfigure}
    
    \vskip\baselineskip
    \begin{subfigure}[c]{1\linewidth}  
        \centering 
        \includegraphics[width=\linewidth]{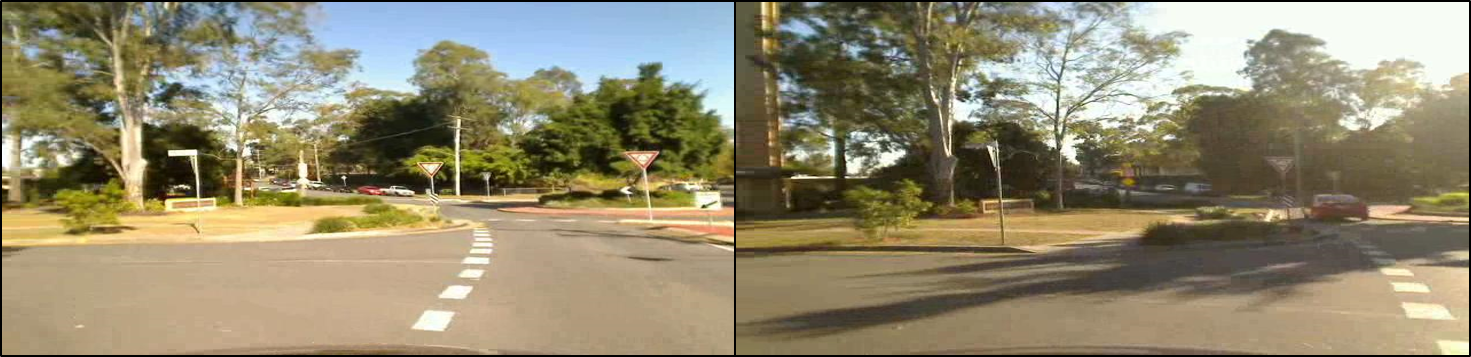}
        \caption{St. Lucia data set: Mild appearance \& severe viewpoint change.}  
        \label{fig:datasets_3}
    \end{subfigure}
    \caption{Appearance and viewpoint variations in data sets on which we test our visual place recognition pipeline.} 
    \label{fig:datasets}
\end{figure}

Robots are often required to operate in environments for a long period of time ranging from days, months to even years. For autonomous operation of robots in such cases, the robot has to recognize places that it has visited before. This ability of the robot to recognize places has a wide range of applications in its autonomous navigation capabilities that include global localization and loop-closure detection. The task of \ac{VPR} for a long-term autonomous visual navigation system becomes extremely challenging because over a long period of time the appearance of a place can drastically change.
Traditional visual place recognition approaches like \cite{fabmap,seqslam} focused on situations where robot has to recognize places that it has recently visited, where the difference between query and database images was mainly due to different view-point of sensors.
However, for a long-term autonomous visual navigation system the \ac{VPR} method should be \textbf{robust} to seasonal, illumination and viewpoint variations (Fig. \ref{fig:datasets}). It is also desired that the VPR system could imbibe robustness to an unfamiliar variation from some learning examples. It should also be \textbf{real-time} \& \textbf{storage efficient} for it to have utility for robotic applications. Additionally, it is advantageous if a visual place recognition system is implementable on a simple hardware configuration so that it can have much wider application. 

Several methods that are robust to certain visual variations have been proposed in the past. S\"{u}nderhauf et al. \cite{gist_closing3} used concatenated BRIEF-gist descriptor to incorporate some robustness to viewpoint variations. Milford et al. \cite{seqslam} proposed to use the video sequence instead of independent images, thereby utilising the continuity constraint of consecutive images in-order to remove outliers.
Lowry et al. \cite{supervised_regression} employed linear regression techniques to predict the temporal appearance of a place based on the time of the day. Neubert et al. \cite{supervised_prediction_superpixel} uses a vocabulary of superpixels to predict the change in appearance of the scene. %Milford et al. \cite{vpr_metric_learning} introduced a more generic supervised visual place recognition method that is capable of learning a variation from its training data. 
%Though, metric learning technique applied in \cite{vpr_metric_learning} is a very slow (often intractable) method of learning. Even after pre-processing using Principal Component Analysis (as suggested in their work), it learns the feature transformation (Mahalanobis) matrix very slowly. 
\newcolumntype{L}{ >{\centering\arraybackslash} m{0.08\linewidth} }
\newcolumntype{M}{ >{\centering\arraybackslash} m{0.12\linewidth} }
\newcolumntype{N}{ >{\centering\arraybackslash} m{0.27\linewidth} }
\newcolumntype{O}{ >{\centering\arraybackslash} m{0.35\linewidth} }

\par
Many visual place recognition methods based on deep \ac{CNN}s which are pre-trained on the task of either object recognition or scene classification, have recently been proposed \cite{vpr_cnn,vpr_cnn_local_regions,mainpaper,rsspaper}.
The deep \ac{CNN} based methods have shown to outperform previous state-of-the-art visual place recognition techniques. However, CNN feature based methods like \cite{vpr_cnn} with no dimensionality reduction are slow because of computations on high dimensional feature vectors. Several dimensionality reduction techniques have been proposed to speed-up the algorithm. Milford et al. \cite{vpr_metric_learning} uses \ac{PCA} as an extra step to reduce the dimensionality of feature vectors. S\"{u}nderhauf et al. \cite{rsspaper} utilized Gaussian random projections to obtain shorter features than the raw {\fontfamily{qcr}\selectfont conv3} \ac{CNN} features. S\"{u}nderhauf et al. \cite{mainpaper} borrows from research in unsupervised hashing methods to obtain binary codes of 8192 bits to describe the images. It is important to note that all the above dimensionality reduction methods are \textbf{unsupervised} and have lower accuracies than a visual place recognition method which uses the `non-reduced' raw features. 
\par 
% VPR approaches based on raw image pixels and simple features like gist or BRIEF \cite{seqslam,fabmap,gist_panorama,gist_closing1,gist_closing2,gist_closing3} involve only basic computations, thus can be easily implemented on simple hardware devices. On the contrary, deep CNN based methods \cite{vpr_cnn,vpr_cnn_local_regions,mainpaper,rsspaper} require hardware compatible to do computations on complex CNN architecture. Moreover, deep CNN architecture itself has millions of parameters, which requires RAM space in the order of hundreds of megabytes. This is difficult to find in simple hardware devices. Also for applications in AI-assisted driving systems, it is injudicious to allocate such large amount of RAM resource to VPR when there are more crucial tasks of obstacle localization, planning etc. in the pipeline, which have much bigger safety concerns.
% \par

Unlike the existing methods, the proposed \ac{VPR} pipeline uses \textbf{supervised} dimensionality reduction to reduce high dimensional real features to compact and more semantic binary codes. 
%The supervised nature of the proposed \ac{VPR} system allows it to learn robustness (from training data) to a variation that image features are otherwise not invariant. 
All other \ac{VPR} methods \cite{vpr_metric_learning, rsspaper, mainpaper} demonstrate lower performance due to the dimensionality reduction step. On the contrary, a supervised \ac{VPR} system based on the learning of compact binary codes from image features significantly improves the image retrieval performance when compared to \ac{VPR} methods based on the corresponding raw features. 
%The proposed methodology exhibits much higher performance on challenging datasets (fig. \ref{fig:datasets}) as compared to pre-trained deep \ac{CNN} based \ac{VPR} methods (unsupervised) \cite{mainpaper,rsspaper}, which are the current state-of-the-art in \ac{VPR} research.
We improve on accuracy while still using a binary code representation. This representation keeps our \ac{VPR} pipeline real-time and space efficient. Also, the proposed method is \textbf{feature agnostic}. Therefore, our \ac{VPR} pipeline is capable of processing both \textit{simple} gist \cite{gist_main} features (whose computation requires only simple convolutions with Gabor filters) as well as \textit{complex} deep learning based CNN features. With the boost of accuracy, our \ac{VPR} method is capable of bootstrapping the accuracies of simple-to-compute features like gist to performance comparable to (usually better than) existing \ac{VPR} methods based on pre-trained CNN features.

% Our works presents a VPR methodology which can incorporate supervision and obtain binary codes in the same computational step. To our knowledge there in no VPR research work which directly leverages \textbf{supervision for learning compact binary codes} from labelled images. Another novelty of our work is that it can be built on simple gist features and still gives better performance than complex CNN based approaches.
\par 
The rest of the paper is organized in the following sections:\\Section \ref{sec:background} describes how gist \& CNN features have been utilized for VPR research. It also motivates the need of a system capable of adaptively \textit{learning} robustness rather than relying only on the \textit{pre-learnt} robustness of popular image descriptors. Section \ref{sec:hashing_methods} describes CCAITQ\cite{itq}, the supervised hashing method that we utilize in our VPR pipeline. Section \ref{sec:experimentation} gives details of the experimentation - data sets, training \& testing set division, assigning similarity labels, running hashing methods and testing. We explain important inferences of our experiments in section \ref{sec:results} and conclude the paper with section \ref{sec:conclusion}.
\section{Background}
\begin{table}[t!]
\label{tab:gist_cnn_compare}
\begin{center}
\begin{tabular*}{1.0\linewidth}{ L M N O }
\hline
\hline
\textbf{Feature} & \textbf{Dimension} & \textbf{Advantage} & \textbf{Disadvantage}\\
\hline
\hline
Gist & 512 or 2048 & Compact global representation & Low performance in severe changes\\
\hline
{\fontfamily{qcr}\selectfont fc6} & 4096 & Robust to viewpoint variations & Susceptible to appearance variations\\
\hline
{\fontfamily{qcr}\selectfont conv3} & 43264 (VGG-f) & Robust to severe appearance variations & Very high dimensional and susceptible to viewpoint variations\\
\hline
\hline
\end{tabular*}
\end{center}
\caption{Details about popular image features for VPR}
\end{table}
\label{sec:background}

\subsection{Gist based Visual Place Recognition methods}
Initial feature based VPR methods (like FAB-MAP \cite{fabmap}) built a visual vocabulary from local SIFT or SURF descriptors and then used a bag-of-words (BoW) image descriptor to find the best matching image frame corresponding to a query frame. Later, success in scene classification based on gist features \cite{gist_bio} led to gist features being applied to the place recognition tasks. Gist features were adapted to panoramic views for VPR \cite{gist_panorama}. Visual Loop Closing methods which earlier utilized SIFT or SURF BoW, demonstrated much superior performance after adapting gist descriptors \cite{gist_closing1,gist_closing2, gist_closing3}. 

\subsection{Deep CNN based VPR}
AlexNet \cite{alexnet} made deep CNNs popular in Computer Vision research in 2012. Studies in \cite{decaf,cnncvprw} demonstrated that features extracted from deep \ac{CNN} (which are pre-trained on object recognition task) can be used for generic visual recognition tasks. CNN features performed better than features which were handcrafted specific for the tasks like domain adaptation, fine grained recognition and scene recognition. Thereafter, research in \ac{VPR} has almost suddenly turned to explore the power of \ac{CNN} based features. Work of S\"{u}nderhauf et al. \cite{mainpaper,rsspaper} has extensively compared features extracted from different \ac{CNN} layers, on challenging VPR data sets. Instead of extracting global CNN image descriptors like \cite{mainpaper}, \cite{rsspaper} extracts pooled local {\fontfamily{qcr}\selectfont conv3} CNN features of fifty landmark regions and then use similarity matching of these local features to obtain the image which is the best match to a query image.
Inferring from the detailed empirical study of \cite{mainpaper}, we focus our experiments on features extracted from two layers~-~ lower level convolutional layer (third layer~-~{\fontfamily{qcr}\selectfont conv3}) and higher level fully connected layer (sixth layer~-~{\fontfamily{qcr}\selectfont fc6}).
% \par\noindent \todo{\textit{Which features to utilize for supervised hashing: Pre-ReLU/Post-ReLU?}\\Since \cite{mainpaper,rsspaper} directly utilize pre-trained CNN features without doing any learning, they use features after ReLU layers (post-ReLU) which are much sparser than pre-ReLU. Study \cite{cnncvprw} has shown that SVM learning over CNN features works better when we use pre-ReLU features. Since our task of supervised hashing method is equivalent to learning multiple (equal to the length of binary codes) hyperplanes, we too use pre-ReLU CNN features in our experiments. We use {\fontfamily{qcr}\selectfont Matconvnet}'s VGG-fast \cite{vgg-f} implementation\footnote{The complete architecture of VGG-fast is available at: \url{http://www.vlfeat.org/matconvnet/models/vgg-face.svg}} which is a minor variant of the {\fontfamily{qcr}\selectfont AlexNet} architecture. The {\fontfamily{qcr}\selectfont conv3} features in VGG-fast are 43,264 dimensional whereas the {\fontfamily{qcr}\selectfont AlexNet}'s {\fontfamily{qcr}\selectfont conv3} features are 64,896 dimensional.}

\par\noindent\\\textbf{Drawbacks of pre-trained CNN based VPR approaches:}
\begin{itemize}
    \item \textit{Choice of layer}: 
    %Different layers have different degree and nature of robustness. 
    \cite{mainpaper} empirically validates the appearance invariance of {\fontfamily{qcr}\selectfont conv3} and and viewpoint invariance of {\fontfamily{qcr}\selectfont fc6} layers. However, for a new environment displaying a unknown weighted mixture of multiple visual variations, one has no insight of which layer features to utilize. 
    %Also, unsupervised approaches can never provide invariance to unseen variations which is desirable in any \ac{VPR} system. Instead, supervised method like ours which are exposed to variations in new (training) data, learns robustness to these variations over the existing pre-trained robustness prevalent in image features.
    
    \item \textit{Dimensionality}: We validated \cite{mainpaper}'s claim that lower depth features from {\fontfamily{qcr}\selectfont conv3} layer are invariant to severe appearance and mild viewpoint variations. Despite their good pre-learnt robustness, raw {\fontfamily{qcr}\selectfont conv3} features are not useful for VPR due to their huge dimensionality, 64896 for \cite{alexnet} and 43264 for \cite{vgg-f}, which significantly increases the computation time. 
    \item \textit{Storage size:} Each dimension for raw real-valued features is 256 bits leading to 1.3Mb size of one {\fontfamily{qcr}\selectfont conv3} feature vector. Table \ref{table-storage} gives details about the storage capacity required for storing different features.
    \item \textit{Hardware requirement}: Deep \ac{CNN}s have complex architecture and require much advanced hardware devices for training and feature extraction. The model of a \ac{CNN} has over millions of parameters and merely loading a \ac{CNN} model requires huge amount of RAM space, which is often impractical for use in many robotic vision applications. 
\end{itemize}
\subsection{Learning invariance vs. pre-learnt invariance}
% As mentioned earlier, the VPR research has recently shifted to pre-trained deep CNN feature based approaches.
Pre-trained CNN features acquire invariance to the variations that the CNN has been exposed to while training. Since the popular pre-trained CNNs are trained for object recognition, the variations in images of the same object might not cover environmental variations prevalent in places. For example, glaring lights in the night traversal of Alderley data set, change of texture from grassy (in summer) to snowy (in winter) are variations which are rarely seen in object recognition data sets. Such variations are not as generic as simple illumination, rotation and scale variations which pre-trained CNNs have already been exposed to, while training on data sets like ImageNet\cite{imagenet}. Thus, any feature based VPR system needs some bootstrapping to learn robustness against \textit{unfamiliar} variations. We incorporate this bootstrapping by supervised hashing methods discussed later. 
\par
\textit{Is obtaining supervision easy:} Obtaining GPS tagged images is cheap and easy, and can be done by any vehicle which travels with a camera and GPS. Multiple traversals of the same tracks helps create cluster of images of the same places at different times of the day and different seasons of the year. This forms the ground truth for a supervised VPR system. There are many publicly available data sets which have multiple images of the same place (under different conditions). These include Nordland \cite{nordland}, Alderley \cite{seqslam}, St. Lucia \cite{st_lucia}, KITTI \cite{kitti}, Pittsburgh 250k \cite{pitts250k} and Tokyo 24/7 \cite{tokyo247} data sets. Thus, supervision is simple to incorporate in VPR and is especially important when a VPR system is put in a new environment.
\begin{table}[t!]

\label{tab:size_features}
\begin{center}
\begin{tabular*}{.98\linewidth}{@{\extracolsep{\fill}}p{0.25\linewidth}p{0.15\linewidth}p{0.13\linewidth}p{0.28\linewidth}@{}}
\hline
\hline
\textbf{Feature} & \textbf{Dimension} & \textbf{Size (bits)} & \textbf{Size of entire Alderley data (MB)}\\
\hline
\hline
Gist512 (binary) & 512 & 512 & 2 Mb\\
\hline
Gist2048 (binary) & 2048 & 2048 & 8 Mb\\
\hline
{\fontfamily{qcr}\selectfont fc6} (binary) & 4096 & 2048 & 8 Mb\\
\hline
Gist512 (raw) & 512 & 131072 & 493 Mb\\
\hline
Gist2048 (raw) & 2048 & 524288 & 1973 Mb\\
\hline
{\fontfamily{qcr}\selectfont fc6} (raw) & 4096 & 1048576 & 3946 Mb\\
\hline
{\fontfamily{qcr}\selectfont conv3} (raw) & 43264 & 11075584 & 41678 Mb\\
\hline
\hline
\end{tabular*}
\end{center}
\caption{Comparing storage size of binary codes and their corresponding raw/precursor features. To illustrate the impact on VPR, we also calculate how much space is required to store a data set (here, Alderley Night/Day data set)}
\label{table-storage}
\end{table}
\section{Hashing methods}
\label{sec:hashing_methods}
In efficient retrieval systems, high dimensional vectors are often compressed using similarity-preserving hash functions which map long features to compact binary codes. Hashing methods can be used to map \textit{similar} images to nearby binary codes (with small hamming distance between them) and map \textit{dissimilar} images to far-away binary codes (with large hamming distance between them). Two main advantages of using hashing methods for visual place recognition are: 
\begin{itemize}
    \item \textit{Storage space}: Table \ref{table-storage} compares the storage size of real-valued features with that of binary code representations. Even though \cite{mainpaper} applies hashing to obtain compact 8192 bit binary codes, it can preserve 95\% of the accuracy obtained by using raw real-valued features. Our supervised approach obtains even shorter binary codes and simultaneously shows better performance than that obtained by using raw real-valued features.
    
    \item \textit{Speed}: The hamming distance computation required to compare similarity between compact binary codes is much faster than euclidean or cosine distance computations required to compare very high dimensional feature vectors. Hamming distances can be efficiently calculated using low level (hardware) bit-wise operations.
\end{itemize}

\par
Hashing methods can be categorized into - unsupervised and supervised, based on the notion of similarity that these methods preserve. Unsupervised hashing methods preserve \textbf{distance based similarity} whereas supervised methods preserve \textbf{label based similarity}.

\subsection{Unsupervised hashing}
Distance based similarity means that images are considered \textit{similar} if their feature vectors have small distance (euclidean, cosine etc.) between them. Unsupervised hashing methods output binary codes which aim to preserve the original distances in real-valued feature space. They leverage no other information except the image features, therefore, their performance is lower than original real-valued features. Despite slightly lower performance, they help overcome two of the issues with CNN features - dimensionality and storage size. Locality sensitive hashing (LSH) \cite{lsh1,lsh2} is a widely used unsupervised hashing method which preserves cosine distance between original data points. Random hyperplane adaptation of LSH in \cite{lsh_charikar} is most commonly used. We too use it in this work to replicate results of previous LSH-based VPR approaches \cite{mainpaper}.
%Unsupervised hashing method are not helpful in scenarios where distances between vectors in image feature space defy the notion of similarity we wish to preserve. To resolve this, we ought to provide labels for learning similarity.

\begin{figure}[t!]
    \centering
    \includegraphics[width=0.8\linewidth]{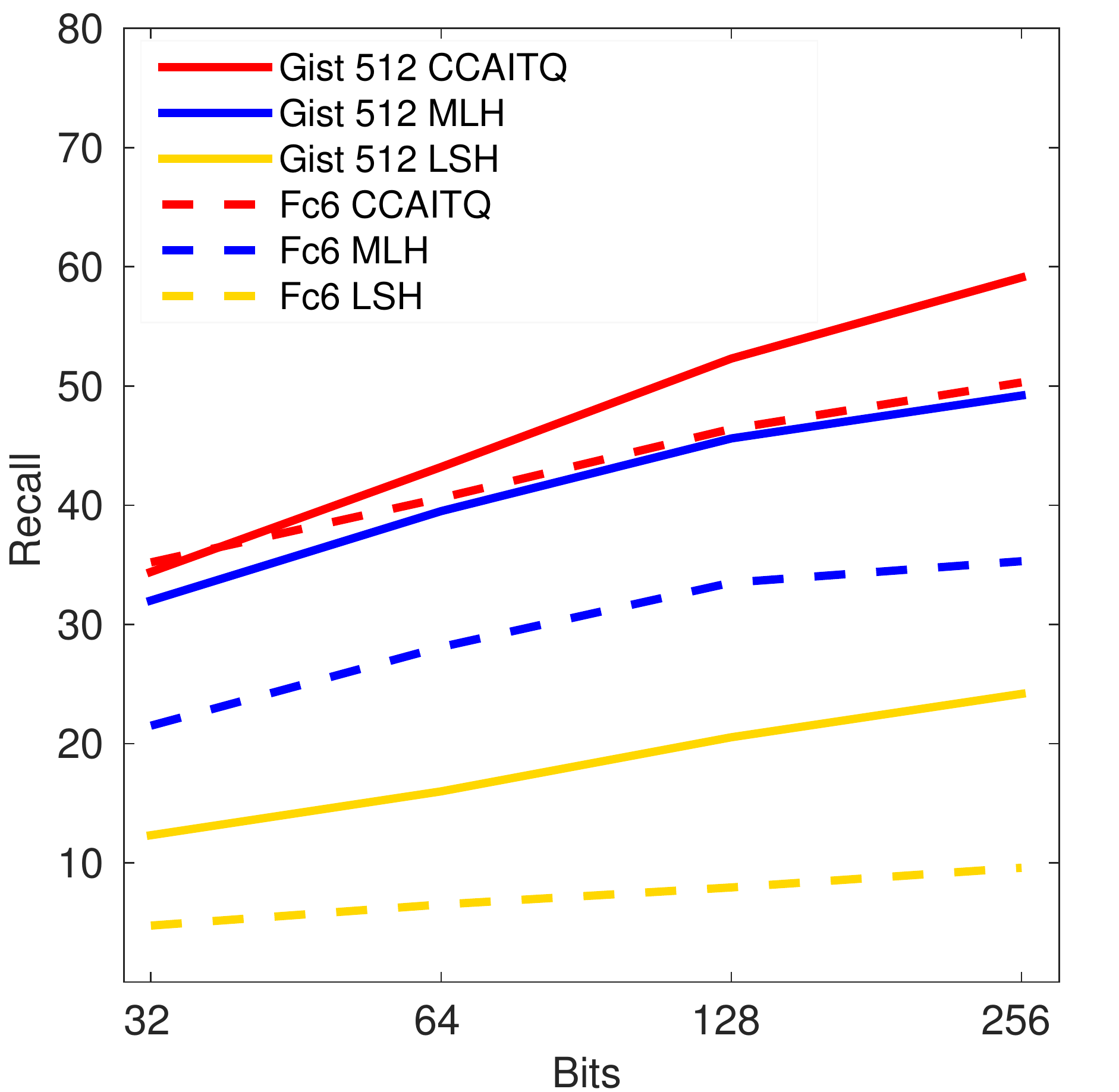}
    \caption{Comparing hashing methods - CCAITQ \cite{itq}, MLH \cite{mlh} and LSH \cite{lsh1}. }
    \label{fig:comparing_MLH}
\end{figure}

\subsection{Supervised hashing}
\label{sec:supervised_hashing}
 Supervised hashing leverage supervision of labels. Using the knowledge from labels, their performance is higher than original real-valued features. For the task of long-term visual place recognition, feature vectors of spatially proximal places might not be nearby in the feature space. For example, feature vector corresponding to winter image of a place X might be closer to the feature vector corresponding to summer image of place Y (rather than that of place X). In such cases externally provided labels can support supervised hashing, which defines a better notion of similarity. We use the well known technique of Canonical Correlation Analysis combined with Iterative Quantization (CCAITQ) \cite{itq} to perform supervised hashing for robust \ac{VPR}. Another supervised hashing technique that was developed at the same time is Minimal Loss Hashing (MLH)~\cite{mlh}. In our comparison in section ~\ref{subsec:comparison} we observe that for the task of VPR, CCAITQ performs better than MLH and is also more computationally efficient. Hence, we use CCAITQ in the proposed VPR method. 
%  but find it to be intractable (more than a day of training time on a 16GB RAM CPU) to learn binary codes on long feature vectors for VPR ({\fontfamily{qcr}\selectfont fc6} and 2048 dimensional gist). Besides, CCAITQ gave better results than MLH on those features where MLH was tractable.

 \par
 CCAITQ \cite{itq} is a linear hashing method which combines Canonical Correlation Analysis (CCA) \cite{cca_original} and Iterative Quantization (ITQ) techniques to eventually obtain binary codes. CCA is the supervised analog of the PCA, which learns a matrix $W \in \mathbb{R}^{d\times k}$ where $d$ is the dimensionality of original features and $k$ is the desired dimensionality of output binary codes. This matrix helps in finding the directions in which the feature vectors and the label vectors have maximum correlation. The input feature matrix is $X = [x_1;x_2;... x_n] \in \mathbb{R}^{n\times d}$ where n is the number of data points and $x_i$'s are rows of features. The aim of CCA is to learn the matrix $W$ such that $V = XW$ transforms feature vectors (rows of $X$) to a more semantic real space. After obtaining transformed representations $v_i\in\mathbb{R}^{k}$ ($i$th row of $V=XW$) from the CCA step, we ought to quantize this representation to obtain binary codes. This can be directly done by using indicator function on each of the dimensions: $f(v)= 1_{\mathbb{R}^{+}}(v)$. However, a better binary embedding 
 is obtained by rotating the features obtained after the CCA step in such a manner that the quantization loss is minimized.
 %would be that which minimizes the quantization loss ($\|f(v)-v\|$) obtained in the second step. This can be achieved by using a different transformation matrix $\widehat{W}=WR$ where $R \in \mathbb{R}^{k \times k}$ is an orthogonal matrix. It can be shown that such a transform of $W$ doesn't alter the CCA formulation which is based on an orthogonal basis of vectors $v_i$. 
 Gong et al. \cite{itq} describe a Procustean approach to solve this quantization problem by minimizing the following loss function:
 $$
    \underset{R}{\arg\min} \|f(VR)-VR\|\text{ s.t. } R'R=RR'=I
 $$
    
By solving this minimization problem we obtain the orthogonal matrix $R$ which minimizes the quantization loss.

\section{Experimentation}
\label{sec:experimentation}
We perform experiments on four data sets chosen for the variety of variation in them (details of data sets is given in table \ref{datasets_detail}). Each data set has two traversals of the same route - database and query traversal, with appropriate ground truth match. In Nordland data set the frames of the database and query traversal are synchronized i.e. $i^{th}$ winter frame matches the $i^{th}$ summer frame, whereas the ground truth in Alderley and St. Lucia data sets is provided externally using a frame matching matrix (\textbf{fm}). More generally, $\textbf{fm}(i)=j$ stores that the $i^{th}$ training frame in query traversal corresponds to the same locations as $j^{th}$ training frame in the database traversal. We use some portion of both the traversals for training CCAITQ to learn the transformation matrices $W$ and $R$ as described in section \ref{sec:supervised_hashing}. 

\begin{figure*}[!t]
    \centering
    \begin{subfigure}[b]{0.32\textwidth}
        \centering
        \includegraphics[width=\textwidth]{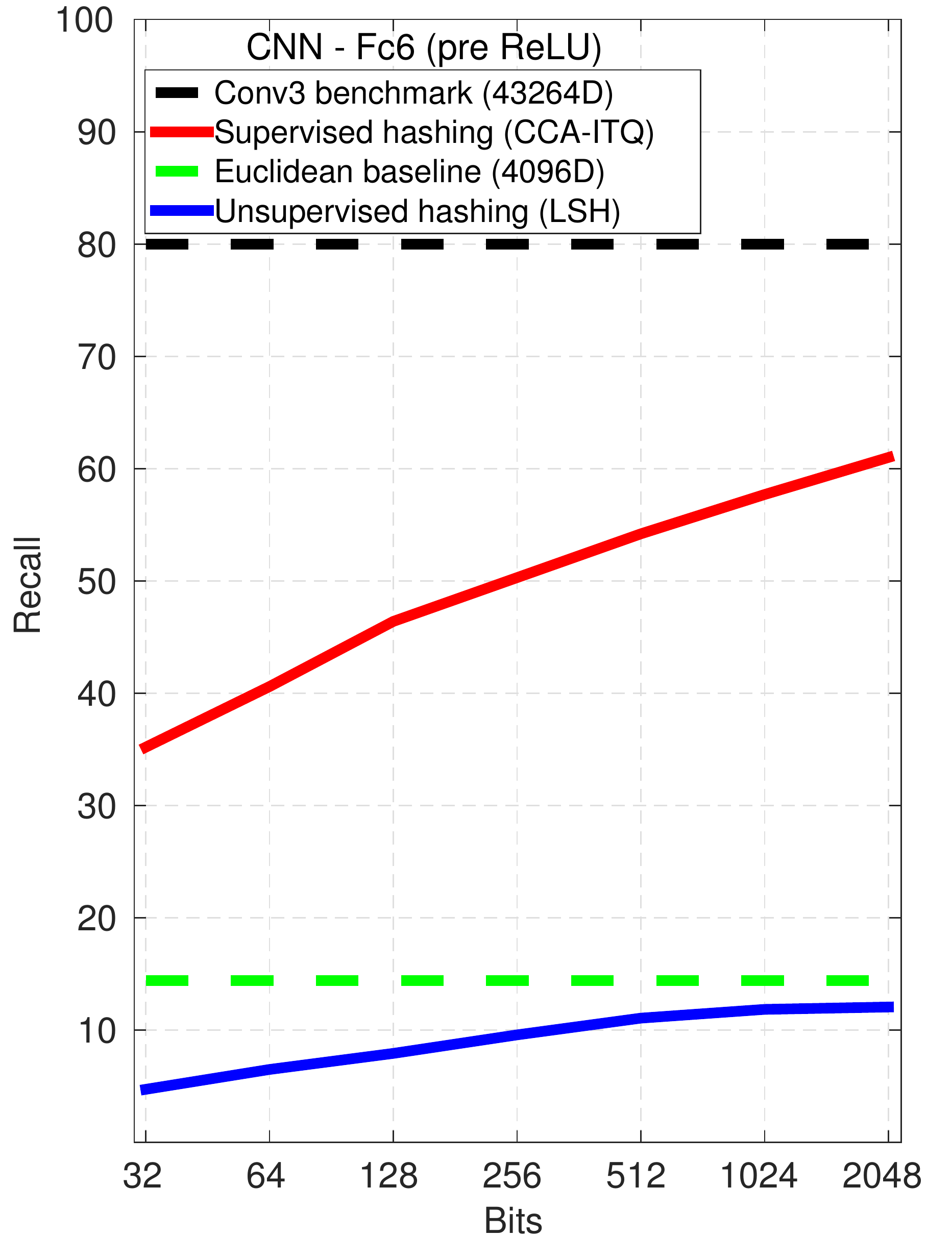}
        \caption{CNN-Fc6 features}    
        \label{fig:bit_comparison1}
        % \label{fig:mean and std of net14}
    \end{subfigure}
    \hfill
    \begin{subfigure}[b]{0.32\textwidth}  
        \centering 
        \includegraphics[width=\textwidth]{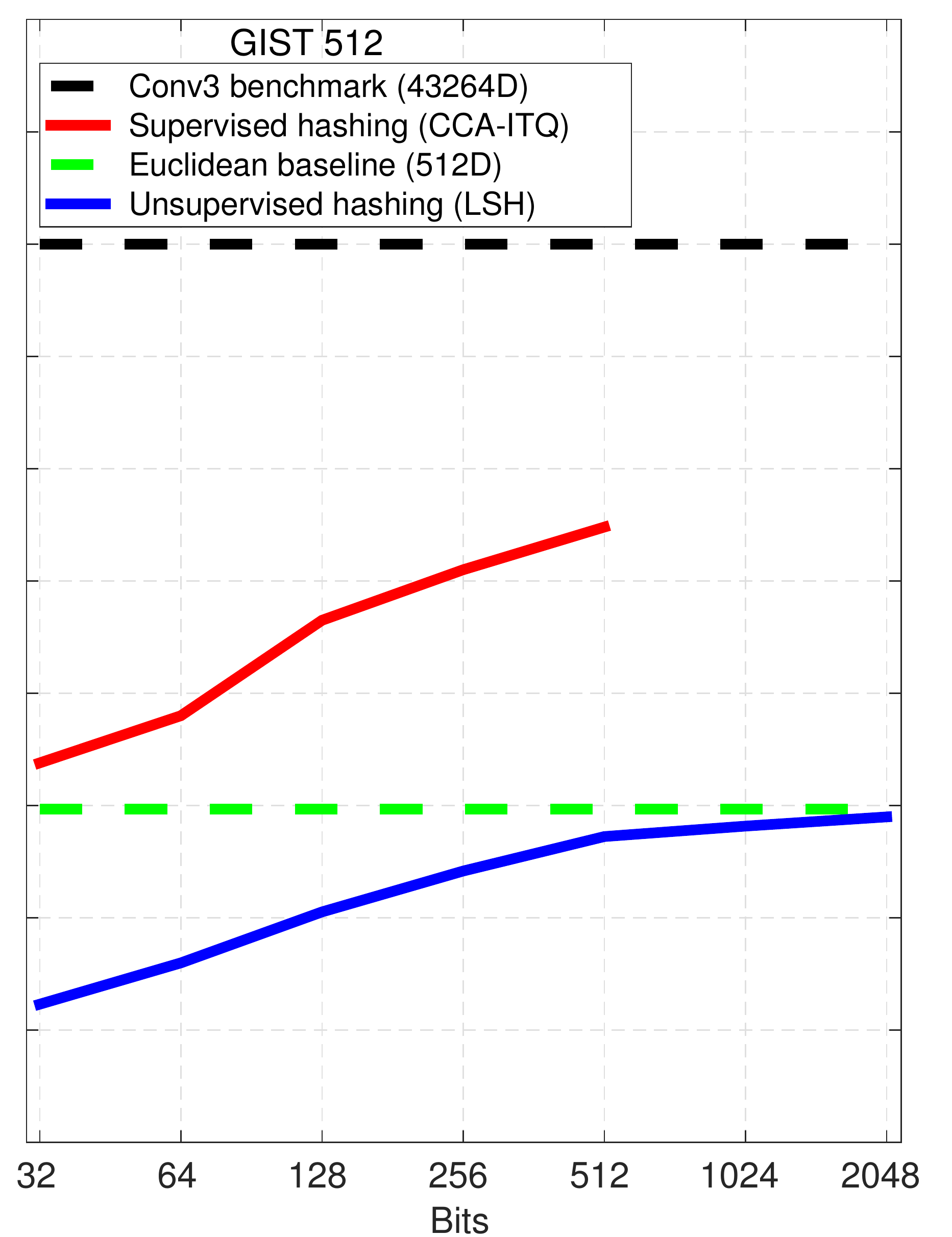}
        \caption{Gist 512D features}  
        \label{fig:bit_comparison2}
    \end{subfigure}
    \hfill
    \begin{subfigure}[b]{0.32\textwidth}  
        \centering 
        \includegraphics[width=\textwidth]{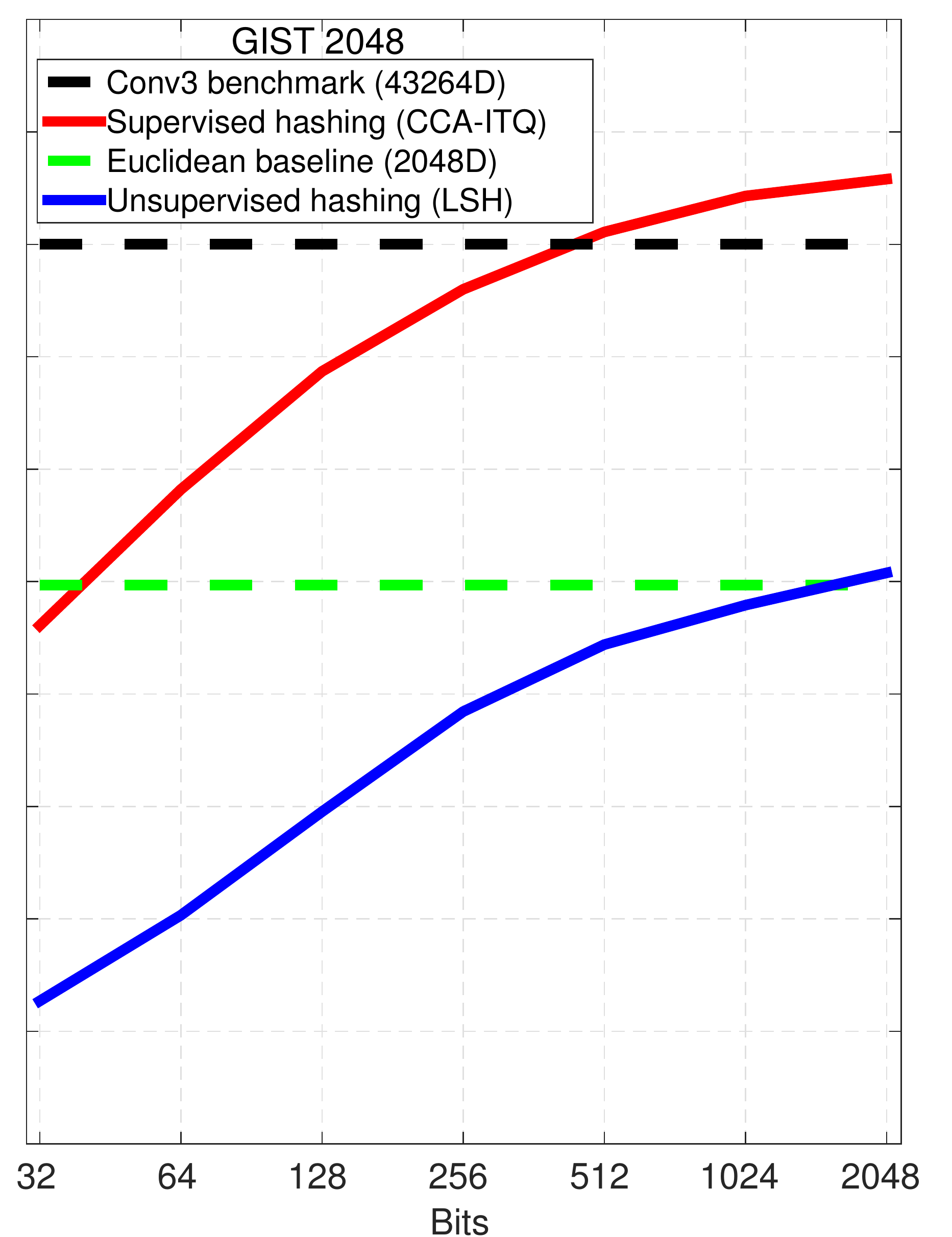}
        \caption{Gist 2048D features}  
        \label{fig:bit_comparison3}
    \end{subfigure}
    \caption{Comparing effect of code length (in bits) on recall (in \%). Our model can convert raw real valued features (\textcolor{green}{green}) to binary codes (\textcolor{red}{red}) which boosts the performance. Binary codes obtained by popular unsupervised hashing method (LSH) (\textcolor{blue}{blue}) have lower performance than the corresponding raw real-valued features.} 
    \label{fig:bit_comparison}
\end{figure*}

\par
For extracting gist feature descriptors we use the original implementation\footnote{\url{http://people.csail.mit.edu/torralba/code/spatialenvelope}} of gist, made available by Oliva \& Torralba. For extracting deep CNN features we use the {\fontfamily{qcr}\selectfont matconvnet} toolbox made available by VGG group.

% We run CCAITQ over pre-ReLU layer features instead of post-ReLU layer features. Also, we use VGG-f \cite{vgg-f}, which is a minor variant of old AlexNet \cite{alexnet} architecture (both are trained on the same ImageNet \cite{imagenet} dataset).

% {\fontfamily{qcr}\selectfont fc6} raw features refers to the post-ReLU $17^{th}$ layer while the {\fontfamily{qcr}\selectfont conv3} benchmark is plotted based on features after the $10^{th}$ layer. The binary codes are obtained after CCAITQ is applied to gist features and to CNN features from the $16^{th}$ layer.
\par
We consider each training image of both traversals as a label. Label vector (for CCAITQ) of a particular training image $i$ has $1's$ for $i\pm~margin$ frames of the same traversal and also $\mathbf{fm}(i)\pm~margin$ frames of the other traversal. Thus, the obtained binary code of a given frame learns similarity to neighbouring frames (\textbf{variation in viewpoint}) and also learn similarity to corresponding frames in the other traversal (\textbf{variation in appearance}). We employ the original implementation of CCA-ITQ\footnote{\url{http://slazebni.cs.illinois.edu/research/smallcode.zip}}. 

%For running comparisons with MLH we use the code\footnote{\url{http://www.cs.toronto.edu/~norouzi/research/mlh/mlh_code_v1.1.tar}} by Narouzi \& Fleet. CCAITQ obtains the binary codes usually in the order of few minutes. On the contrary, MLH is intractable (requires more than a day for training) for binary codes beyond 256 bits.

\subsection{Comparing supervised hashing methods}
\label{subsec:comparison}
The works of MLH \cite{mlh} and CCAITQ \cite{itq} have performed well in comparison to other methods of supervised hashing such as spectral hashing and binary reconstructive embedding. We therefore restrict ourselves in this paper to these two popularly used supervised hashing methods. 
Fig. \ref{fig:comparing_MLH} shows a comparison between the two hashing methods. We also compare it to the baseline of unsupervised hashing method LSH (used in recent VPR methods \cite{mainpaper, rsspaper}). We find MLH is intractable for learning more than 256 bit binary codes, requiring more than a day of training time on our data sets on a 16Gb i7 computer. On the contrary, CCAITQ requires only few minutes to train over 4096D {\fontfamily{qcr}\selectfont fc6} features for data sets of considerable size. Moreover, it always outperforms MLH's performance. Hence, we conduct experiments and report results using CCAITQ in our supervised VPR pipeline.

\subsection{Bit-wise study of supervised VPR}
We compare the recall performance of binary codes for different code lengths varying from 32 to 2048 bits for the most challenging Nordland winter-summer data set. While testing our VPR pipeline, we extract binary codes for each test set image of query and database traversal (using learnt $W$ and $R$ matrices). For every query image code we find the closest database image code (in hamming space) and count it a true positive if it is within the \textit{margin} around ground truth match. CCAITQ algorithm outputs binary codes with lengths lesser than the dimensionality of the raw features. Hence, \ref{fig:bit_comparison2} does not go beyond 512 bits (learnt from 512D gist features). The \textcolor{black}{black} benchmark is calculated using {\fontfamily{qcr}\selectfont conv3} raw features, which is the best performing feature as suggested in \cite{mainpaper}. We compare {\fontfamily{qcr}\selectfont conv3} features with our VPR pipeline's results, obtained from much smaller features~-~simple gist and {\fontfamily{qcr}\selectfont fc6} CNN features. We observed that the learnt binary codes (\textcolor{red}{red}) perform significantly better than the corresponding raw features (\textcolor{green}{green}). Fig. \ref{fig:bit_comparison3} shows the VPR system with the proposed modification helps outperforms {\fontfamily{qcr}\selectfont conv3} raw features \cite{mainpaper} by using only 2048 bit binary codes. Hence we are able to \textbf{bootstrap simple-to-compute \& low dimensional gist features to match the performance of pre-trained CNN based VPR systems} and \textbf{\textit{simultaneously} reduce dimensionality \& storage space}.

\begin{figure*}[t!]
    \centering
    \begin{subfigure}[b]{0.49\textwidth}
        \centering
        \includegraphics[width=1.0\linewidth]{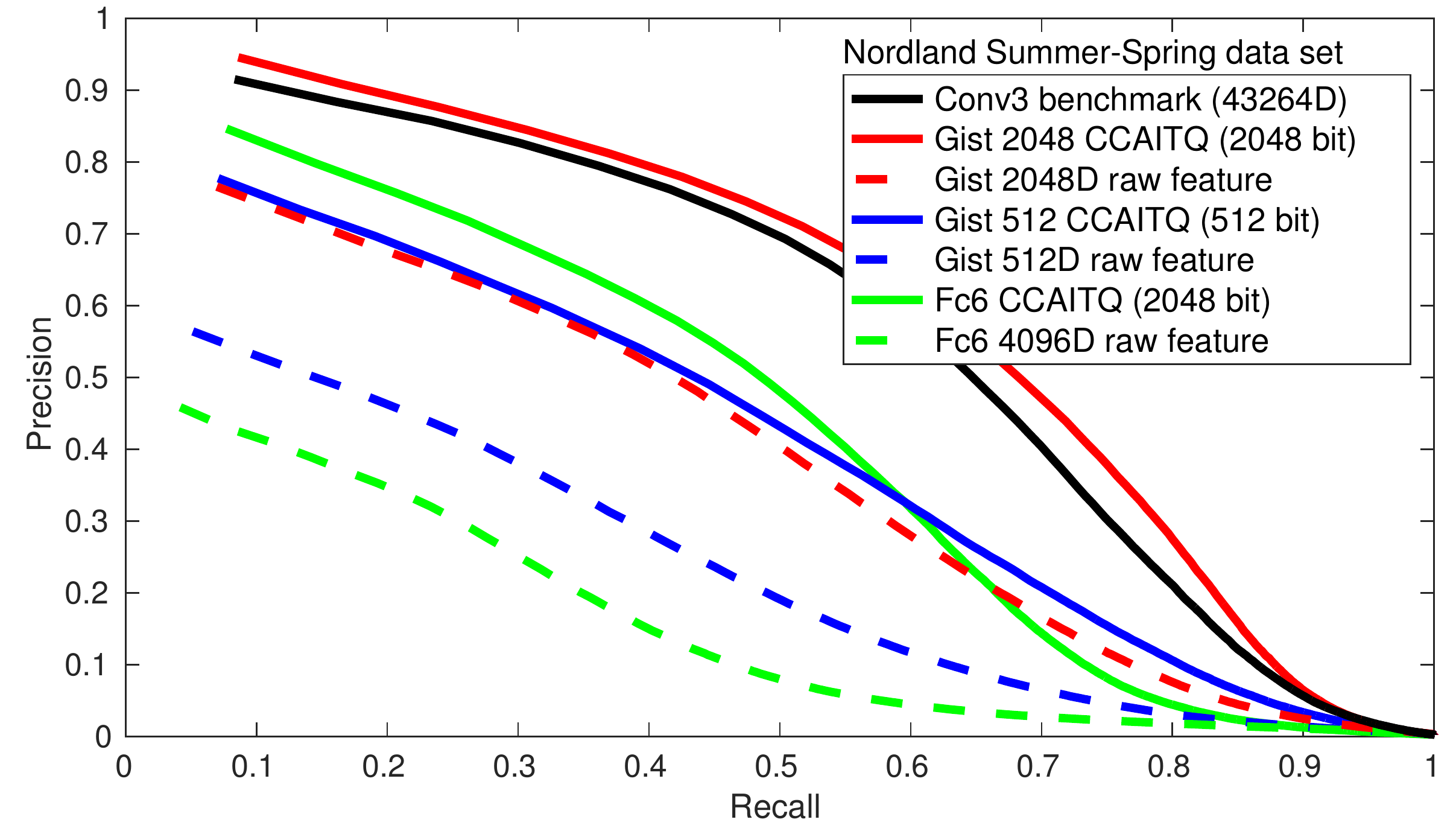}
        \subcaption{Nordland summer-spring data set (A: Mild, V: No)}
        \label{fig:NSS_final}% \label{fig:mean and std of net14}
    \end{subfigure}
    \hfill
    \begin{subfigure}[b]{0.49\textwidth}  
        \centering
        \includegraphics[width=1.0\linewidth]{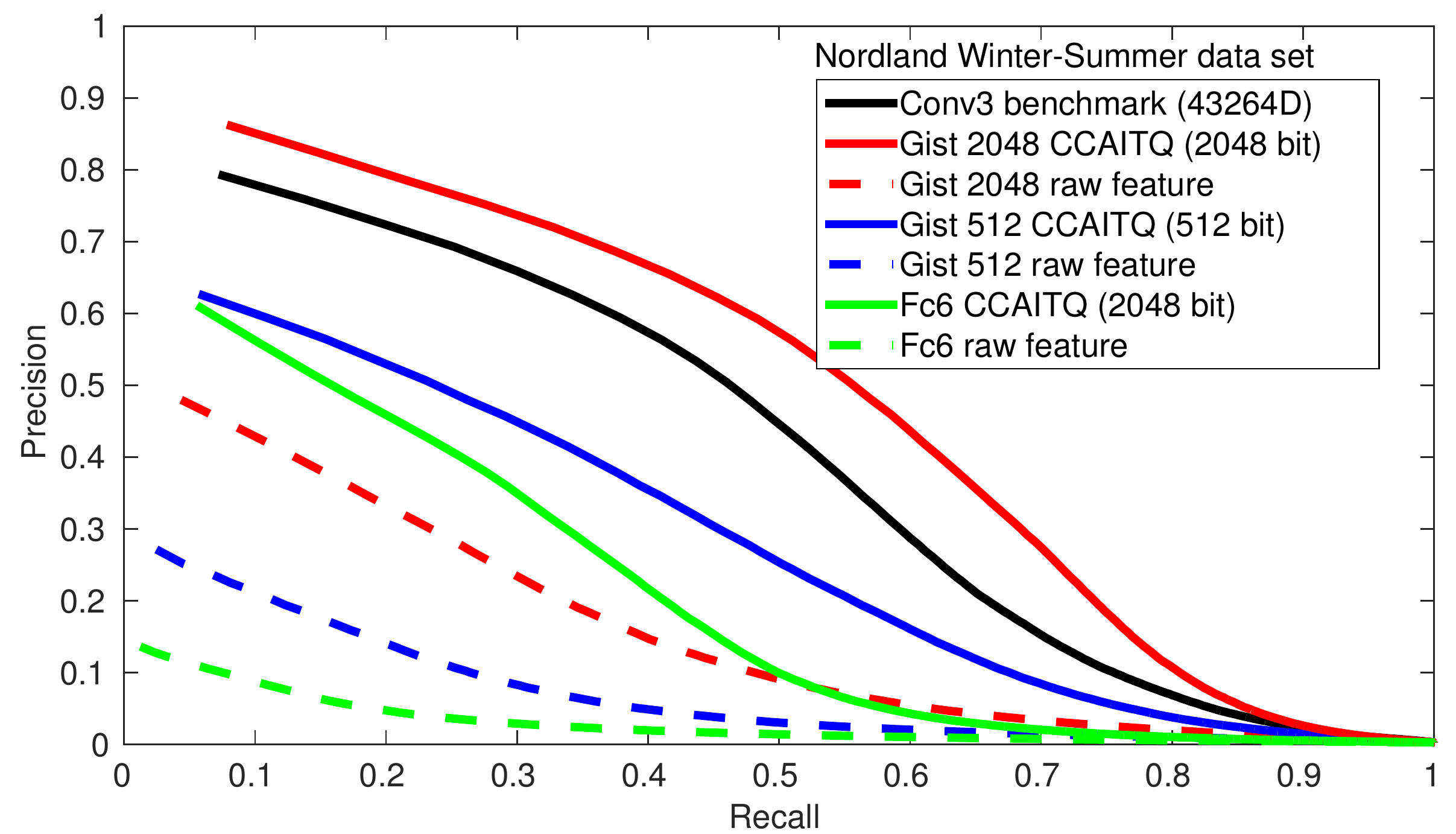}
        \subcaption{Nordland summer-winter data set (A: Severe, V: None)}
        \label{fig:NWS_final}
    \end{subfigure}
    \\
    \begin{subfigure}[b]{0.49\textwidth}  
        \centering
        \includegraphics[width=1.0\linewidth]{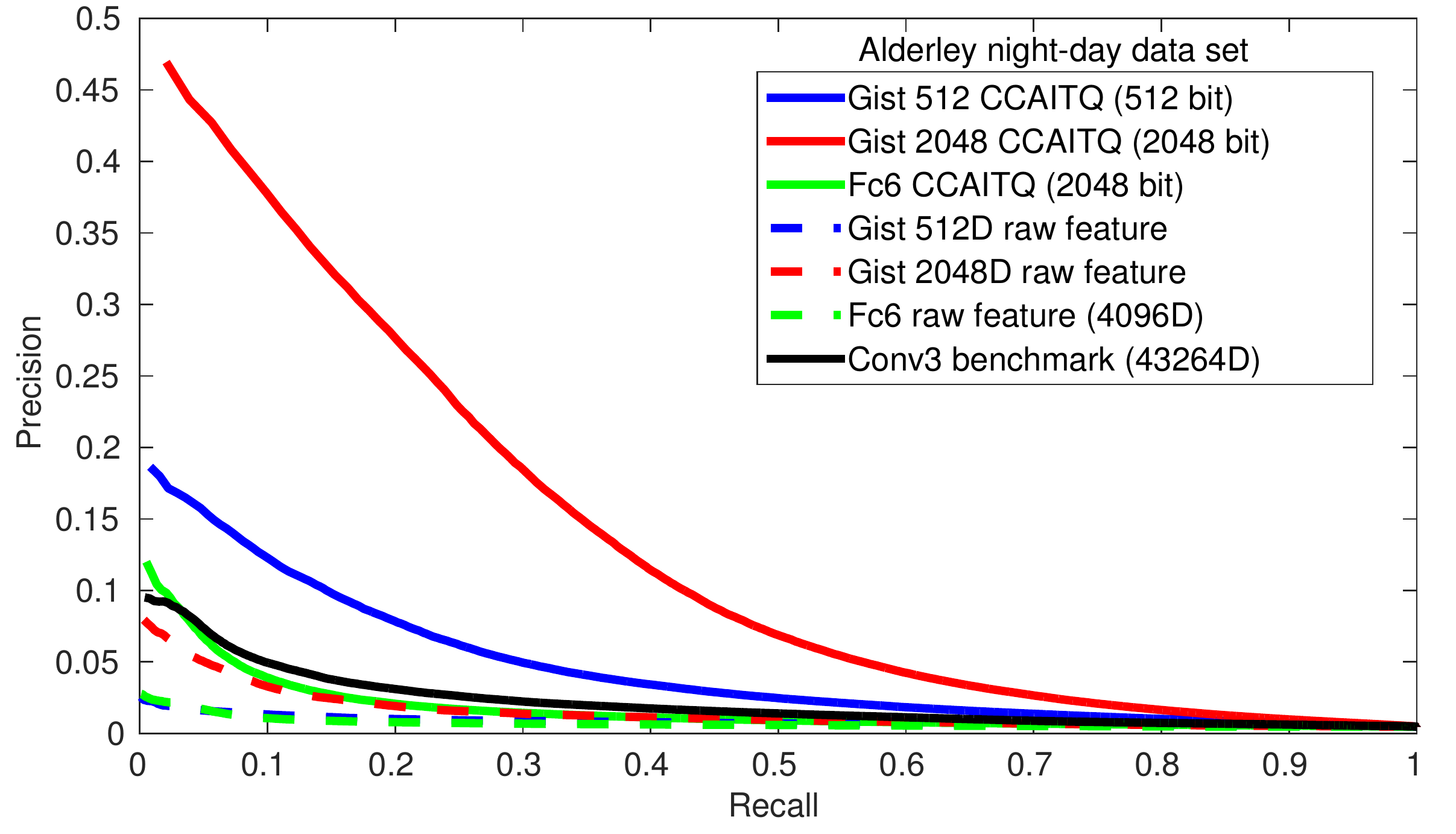}
        \subcaption{Alderley Night-Day data set (A: Severe, V: Mild)}
        \label{fig:alderley_final}
    \end{subfigure}
    \hfill
    \begin{subfigure}[b]{0.49\textwidth}  
        \centering
        \includegraphics[width=1.0\linewidth]{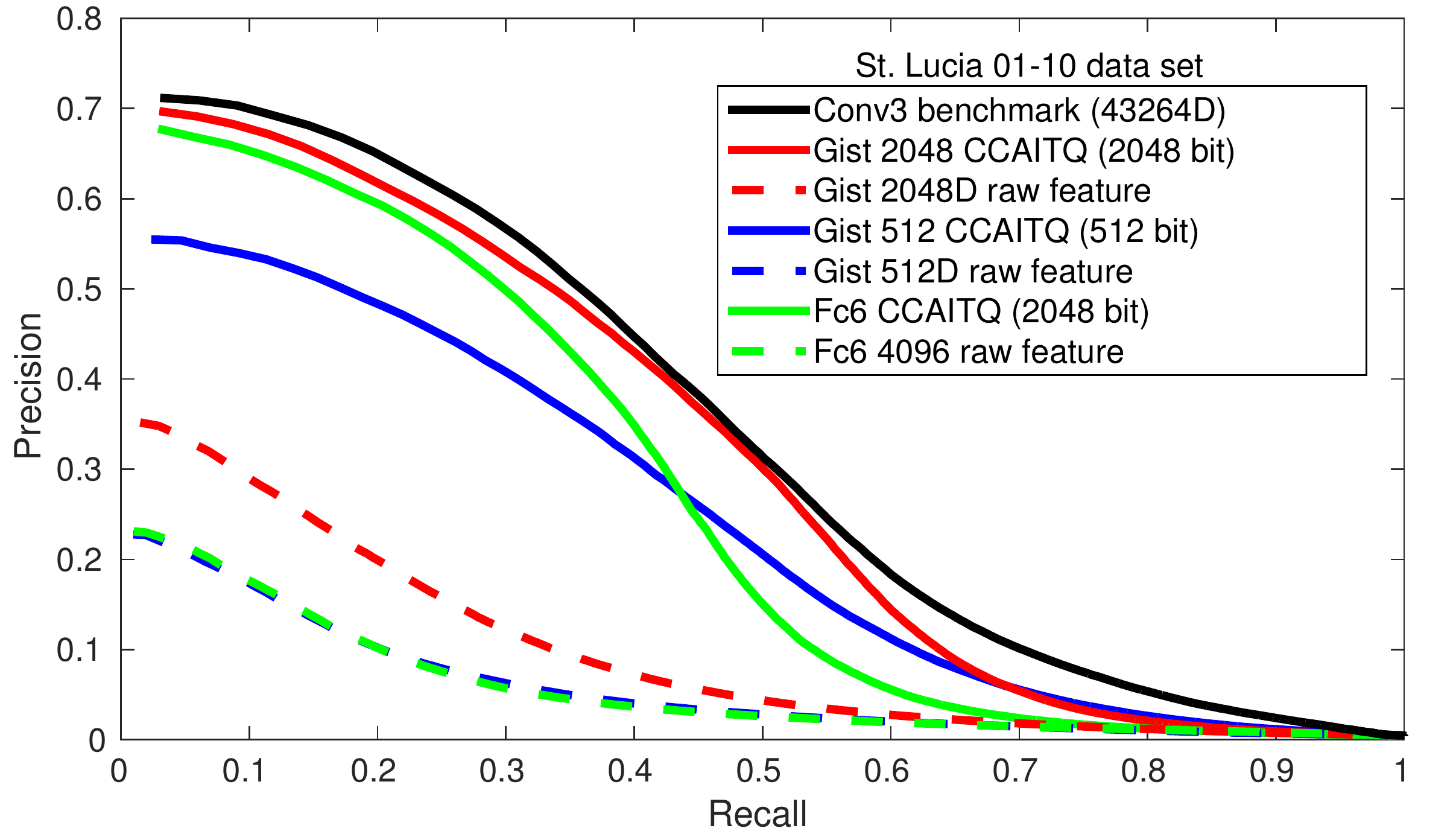}
        \subcaption{St. Lucia 01-10 data set (A: Mild, V: Severe)}
        \label{fig:stlucia_final}
    \end{subfigure}
    \caption{Precision-Recall curves of data sets exhibiting difference mixture of appearance (A) and viewpoint (V) variations} 
    \label{fig:PR-curve}
\end{figure*}

\subsection{Comparing precision-recall performance}
\label{sec:results}
Precision-Recall (PR) curves are used to compare image retrieval techniques. VPR is similar to image retrieval and research in VPR \cite{mainpaper,rsspaper,vpr_cnn,vpr_metric_learning,seqslam} plot PR curves by altering a parameter. Achieving high precision with high recall is desired. Hence, farther a PR plot is from the axes, better the performance. The procedure for plotting PR curve for a VPR system is described ahead. Each query image has $n=2*margin+1$ ground truth positives. We retrieve top $k$ matches for a query image out of which $m$ ($\leq k$) are true positives. We use $Precision=m/k$ and $Recall=m/n$, which is averaged over all query images to make the PR plot. The value of $k$ is varied from 1 to the total number of test images to obtain multiple points on the curve. We extract different top matches for each VPR method and compare the performance in fig. \ref{fig:PR-curve}. We observe that the recognition performance of the binary codes (coloured solid lines) learnt over all three features - \textcolor{blue}{512D gist}, \textcolor{red}{2048D gist} and \textcolor{green}{4096D {\fontfamily{qcr}\selectfont fc6}}, improve over their corresponding raw feature versions (coloured dashed lines). Moreover, \textbf{2048 bit binary codes learnt over 2048D gist features shows better (or marginally less) performance than the benchmarking raw {\fontfamily{qcr}\selectfont conv3} features} (solid \textbf{black}).

\subsection{Leveraging continuity information using contrast enhanced Dynamic Time Warping}

\begin{figure}[t!]
    \centering
    \includegraphics[width=1.0\linewidth]{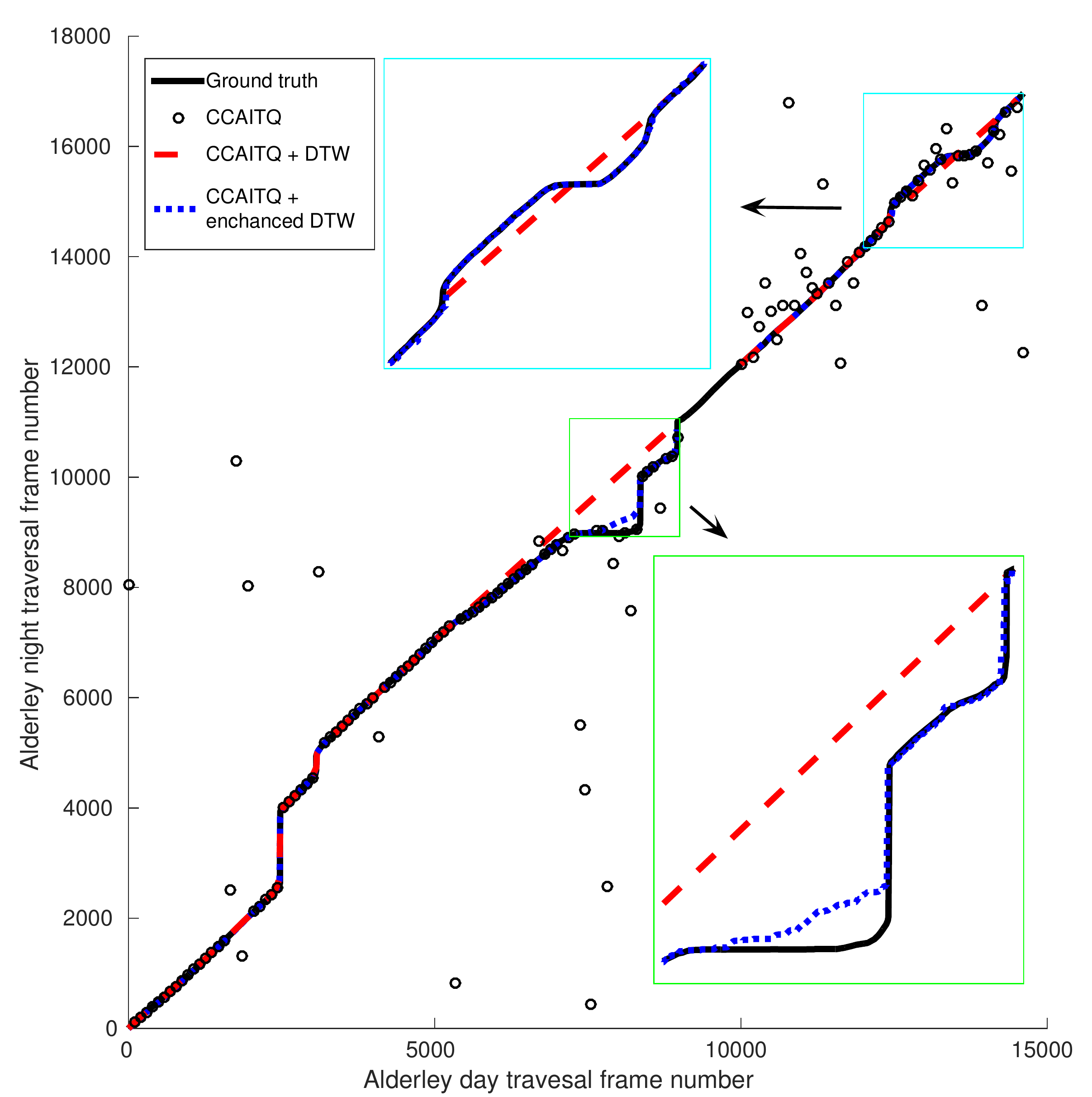}
    \caption{Impact of leveraging sequencing or continuity information for Alderley data set.}
    \label{fig:sequencing}
\end{figure}

While most VPR methods are similar to an image retrieval for each query image, some methods like \cite{seqslam,smart} leverage the fact that we always have a sequence of images as opposed to a single query image. The common assumption being that the two traversals have no negative velocities (reverse/backward travel). Authors of \cite{seqslam} suggest using Dynamic Time Warping (DTW) \cite{dtw} in future work to tackle variations in velocities. We apply DTW over our binary codes (\textcolor{red}{red} dashed) to show an improvement in results. The results are compiled in fig. \ref{fig:sequencing} where the individual image-level VPR method (black circles) explained till now gives slightly incongruous (non continuous) retrievals. Cost of a path has two factors - number of elements (length of path) and cost of each element (divergence from ground truth). We contrast enhanced the DTW distance matrix, thus, giving more weight to the factor of divergence from ground truth. Hence, we found such a cost function (\textcolor{blue}{blue} dots) to give better performance on non-diagonal trajectories (zoomed in plots in fig. \ref{fig:sequencing}).

\newcolumntype{C}{ >{\centering\arraybackslash} m{0.06\linewidth} }
\newcolumntype{D}{ >{\centering\arraybackslash} m{0.11\linewidth} }
\newcolumntype{E}{ >{\centering\arraybackslash} m{0.15\linewidth} }
\newcolumntype{F}{ >{\centering\arraybackslash} m{0.15\linewidth} }
\newcolumntype{G}{ >{\centering\arraybackslash} m{0.08\linewidth} }
\newcolumntype{H}{ >{\centering\arraybackslash} m{0.08\linewidth} }
\begin{table*}[t!]
\begin{center}
\begin{tabular*}{\linewidth}{C H H D E F G G} 
%{@{\extracolsep{\fill}}p{0.06\linewidth}p{0.06\linewidth}p{0.06\linewidth}p{0.20\linewidth}p{0.1\linewidth}p{0.13\linewidth}p{0.03\linewidth}p{0.03\linewidth}@{}}
\hline
\hline
\textbf{Data set} & \textbf{Database traversal} & \textbf{Query traversal} & \textbf{Variation present} & \textbf{Train frames} & \textbf{Test frames} & \textbf{Sample rate (fps)} & \textbf{Margin ($\pm$~frames)}\\
\hline
\hline
Nordland & Summer & Spring & A:~Mild, V:~None & $1-10k$ (both~traversals) & $11k-16k$ (both~traversals) & 2 & 5\\
\hline
Nordland & Summer & Winter & A:~Severe, V:~None & $1-10k$ (both~traversals) & $11k-16k$ (both~traversals) & 2 & 5\\
\hline
Alderley & Night & Day & A:~Severe, V:~Mild & $1-9k$ (day) $1-11301$ (night) & $10k-14607$ (day) $12051-16960$ (night) & 25 & 10\\
\hline
St. Lucia & $1^{st}$ traversal & $10^{th}$ traversal & A:~Mild, V:~Severe & $1-10k$ (10) $1-11145$ (01) & $11001-16k$ (10) $12145-17512$ (01) & 15 & 10\\
\hline
\hline
\end{tabular*}
\end{center}
\caption{Experimental details and variations present in data sets used for evaluation}
\label{datasets_detail}
\end{table*}

\subsection{Implementation details}
Additional details of implementation needing explanation are as follows:
\begin{itemize}
    \item \textit{CCAITQ is applied over pre-ReLU layer features instead of post-ReLU layer features:} Rectified linear unit (ReLU) layer: $f(x)=max(x,0)$ add the non-linearity quotient to deep CNNs. Since \cite{mainpaper,rsspaper} directly utilize pre-trained CNN features without doing any learning, they may choose to use features after ReLU layers (post-ReLU) which are much sparser than pre-ReLU. Studies in \cite{cnncvprw} have shown that SVM learning over CNN features works better when we use pre-ReLU features. Since our task of supervised hashing method is equivalent to learning multiple (equal to the length of binary codes) hyperplanes, we too use pre-ReLU CNN features in our experiments. For features requiring no learning, we report results using post-ReLU (same as \cite{mainpaper, rsspaper}).
    \item \textit{Use of VGG-f:} VGG-f \cite{vgg-f} has five convolutional, three fully connected layers and takes a $224 \times 224$ image (all same as AlexNet\cite{alexnet}). While the old AlexNet architecture is configured rigidly to accommodate training distributed over two GPUs, VGG-f gets rid of such dependency. With more minor variations, it is shown to give better performance than AlexNet on recognition tasks. Hence, we report results using this variant and achieve better performance for both raw features as well as binary codes.
    \item \textit{Supervised hashing not applied on {\fontfamily{qcr}\selectfont conv3} layer:} Whether we choose VGG-f or AlexNet variant of the deep CNN, the dimensionality of {\fontfamily{qcr}\selectfont conv3} layer is too high to tractably run CCAITQ (MLH is even slower). Both aims of our VPR pipeline - Speed and storage efficiency (Sec. \ref{sec:hashing_methods}), are incongruent to the use of {\fontfamily{qcr}\selectfont conv3} layer features. Supervised hashed codes that we obtain from gist or {\fontfamily{qcr}\selectfont fc6} CNN features are compared to the raw {\fontfamily{qcr}\selectfont conv3} layer (black plots in fig. \ref{fig:bit_comparison} and \ref{fig:PR-curve}).
    
\end{itemize}

\section{Conclusion}
\label{sec:conclusion}
To the best of our knowledge, our work is the first to introduce the successful learning methods of supervised hashing to the Visual Place Recognition research. The application is a perfect fit as there is a need for both \textit{learning} and \textit{compact representation} for any VPR to be robust and real-time, respectively. There is the alternative to learn these compact embeddings by training a CNN for this task. Training of CNNs require a long time and also require much complex hardware capable systems. We conclude that for a widespread application of VPR, supervised hashing is ideal as it is both quick to train and outputs compact binary representation of images.
\section*{Acknowledgment}
The authors would like to thank Ford Motor Company for supporting this research through the project FMT/EE/2015241



%\bibliographystyle{IEEEtran}
\bibliographystyle{IEEEtran}
\bibliography{references}


% \end{thebibliography}

\end{document}